%% file: smoothness.tex
\documentclass[abbrvbib]{article}
\usepackage[numbers,sort]{natbib}
\usepackage[]{amsfonts}
\usepackage{amsmath}

\usepackage[dvipsnames]{xcolor}
\usepackage[colorinlistoftodos]{todonotes}
\usepackage{caption}
\usepackage{graphicx}
\usepackage{subcaption}
\usepackage{coseoul}
\usepackage{fixltx2e} %

\usepackage[nonatbib,final]{ICBINB_latex_style_template/icbinb_neurips_2020}

\usepackage[utf8]{inputenc} %
\usepackage[T1]{fontenc}    %
\usepackage{hyperref}       %
\usepackage{url}            %
\usepackage{booktabs}       %
\usepackage{amsfonts}       %
\usepackage{nicefrac}       %
\usepackage{microtype}      %

\usepackage[justification=centering]{caption}

\input{macros}

\title{A case for new neural network smoothness constraints}

\author{
 Mihaela Rosca$^{1, 2}$ \quad
Theophane Weber$^1$ \quad
 Arthur Gretton$^2$ \quad
Shakir Mohamed$^1$\\
  $^1$DeepMind \hspace{4em}  $^2$University College London\\
  \texttt{\{mihaelacr,theophane,shakir\}@google.com, arthur.gretton@gmail.com}
}

\begin{document}

\maketitle

\begin{abstract}

How sensitive should machine learning models be to \textit{input} changes?
We tackle the question of  model smoothness and show that it is a useful inductive bias which aids generalization, adversarial robustness,
generative modeling and reinforcement learning.
We explore current methods of imposing smoothness constraints and observe they
lack the flexibility to adapt to new tasks,
they don't account for data modalities,
they interact with losses, architectures and optimization in ways not yet fully understood.
 We conclude that new advances in the field are hinging on finding ways to
 incorporate \textit{data}, \textit{tasks} and \textit{learning}
 into our definitions of smoothness.
\end{abstract}

\section{Introduction}

How certain should a classifier be when it is presented with out of distribution data?
How much mass should a generative model assign around a datapoint?
How much should an agent's behavior change when its environment changes slightly?
Answering these questions shows the need to quantify the manner in which the output of a
function varies with changes in its input, a quantity we will
intuitively call the smoothness of the model.
Today, when learning a smooth function using neural networks the machine learning practitioner
 is bound to choose between regularization techniques whose effect
on smoothness is poorly understood and rigid techniques that do not account for the data or task at hand.
Despite these shortcomings, imposing smoothness constraints on neural networks  has led to great progress
in machine learning, from boosting generalization and robustness of classifiers to
increasing the stability and performance of generative models
and providing better priors for reinforcement learning agents.
We use the potential of smoothness and the downsides of current approaches
to construct the case for incorporating tasks, data modality and learning into smoothness definitions
and argue for a more integrated view of smoothness constraints and their interaction with losses, models and optimization.

This paper discusses the smoothness of functions parametrized using neural networks;
other classes of functions such as reproducing kernel Hilbert spaces
use different notions of smoothness which are outside our scope. Inside the neural network family of functions, we are looking at model smoothness with respect to \textit{inputs}; we do not consider smoothness with respect to parameters.

\section{Measuring function smoothness}

Neural network ``smoothness'' is a broad, vague, catch all term.
 We use it to convey formal definitions such as differentiable, bounded, Lipschitz,
 as well as intuitive concepts such as invariant to data dimensions or projections, robust to input perturbations, and others.
One definition states that a function $f: \mathcal{X} \rightarrow \mathcal{Y}$ is $n$ smooth if is $n$ times differentiable with the $n$-th derivative being continuous.
  The differentiability of a function
 is not a very useful inductive bias for a model, as it is both very local and constructed according to the metric of the space where limits are taken.
 What we are looking for is the ability to choose both the distance metric and how local or global our smoothness inductive biases are.
 With this in mind,
 Lipschitz continuity is appealing as it defines a \textit{global} property and provides the choice of distances in the domain and co-domain of $f$. It is defined as:
\begin{align}
  \norm{f(\vx_1) - f(\vx_2)}_{\mathcal{Y}} \le K \norm{\vx_1-\vx_2}_{\mathcal{X}}  \hspace{3em} \forall \vx_1, \vx_2 \in \mathcal{X}
  \label{eq:lip_def}
\end{align}

where $K$ is denoted as the Lipschitz constant of function $f$.
Enforcing Equation~\ref{eq:lip_def} can be difficult, but according to Rademacher's theorem if $\mathcal{X} \subset \mathbb{R}^m$ is an open set and $\mathcal{Y} = \mathbb{R}^p$ and $f$ is $K$-Lipschitz then $\norm{D f(\vx)} \le K$ wherever the total derivative $D f(\vx)$ exists.
For $p=1$, this entails $\norm{\nabla_\vx f(\vx)} \le K$ wherever $f$ is differentiable.
Conversely, a function that is differentiable everywhere with bounded gradient norm is Lipschitz.
Thus, a convenient strategy to make a differentiable function $K$-Lipschitz is to ensure $\norm{\nabla_\vx f(\vx)} \le K, \forall \vx \in \mathcal{X}$.

If $f$ and $g$ are Lipschitz with constants $K_f$ and $K_g$, $f \circ g$ is Lipschitz with constant $K_f K_g$. Since commonly used activation functions are 1-Lipschitz, the task of ensuring a neural network is Lipschitz reduces to constraining the learneable layers to be Lipschitz. Many neural networks layers are linear operators (linear and convolutional layers, BatchNormalization~\citep{batch_norm}), and to compute their Lipschitz constant we can use that the Lipschitz constant of a linear operator $A$ under common norms such as $l_1, l_2, l_{\infty}$ is $\sup_{\vx \ne 0} \frac{\norm{A\vx}}{\norm{\vx}}$.

To avoid learning trivially smooth functions and maintain useful variability, it is often beneficial to constrain the function variation both from above and below. This leads to bi-Lipschitz continuity:
\begin{align}
  K_1 \norm{\vx_1-\vx_2}_{\mathcal{X}} \le \norm{f(\vx_1) - f(\vx_2)}_{\mathcal{Y}} \le K_2 \norm{\vx_1-\vx_2}_{\mathcal{X}}
\end{align}

Another way to measure smoothness
is through various matrix norms of the Jacobian $J(\vx) = \frac{d f(\vx)}{d \vx}$.
Instead of constraining the total derivative as in Lipschitz continuity, Jacobian metrics
account for how \textit{each dimension} of the function output is allowed to vary as individual input dimensions change.

\section{Smoothness regularization for neural networks}
\label{sec:smoothness_techniques}

Smoothness regularizers have long been part of the toolkit of the machine learning practitioner:
early stopping encourages smoothness by stopping optimization before the model overfits the training data;
 dropout~\citep{dropout} makes the network more robust to small changes in the input by randomly masking hidden activations; max pooling encourages smoothness with respect to local changes;
 $L_2$ weight regularization and weight decay~\citep{loshchilov2017decoupled} discourage large changes in output by not allowing individual weight norms to grow;
 data augmentation allows us to specify what changes in the input should not result in large changes in the model prediction and thus is also closely related to smoothness and invariance to input transformations.
  These smoothness regularization techniques are often introduced as methods which directly target generalization and other beneficial effects of smoothness discussed in Section~\ref{sec:benefits}, instead of being seen through the lens of smoothness regularization.

Methods which explicitly target smoothness on the entire input space focus on restricting the learned model family.
A common approach is to ensure Lipschitz smoothness with respect to the $l_2$ metric by individually restricting each layer to be Lipschitz.
Spectral regularization~\citep{yoshida2017spectral} uses the sum of the spectral norms -  the largest singular value - of each layer as a regularization loss to encourage Lipschitz smoothness. Spectral Normalization~\citep{miyato2018spectral} ensures the learned models are 1-Lipschitz by adding a node in the computational graph of the model layers by replacing the weights with their normalized version: $\mathcal{L}(W)$ becomes $\mathcal{L}(\sigma(W))$, where $\sigma(W) = W / ||W||_2$  and $||W||_2$ is the spectral norm of $W$. Both methods use power iteration to compute the spectral norm of weight matrices.~\citet{gouk2018regularisation} use a projection method by dividing the weights by the spectral norm after a gradient update. This is unlike Spectral Normalization, which backpropgates through the normalization operation.
The majority of this line of work has focused on constraints for linear and convolutional layers, and only recently attempts to expand to other layers, such as self attention have been made~\citep{lipschitz_constant_self_attention}.
Efficiency is always a concern and heuristics are often used even for popular layers such as convolutional layers~\citep{miyato2018spectral} despite more accurate algorithms being available~\citep{gouk2018regularisation,tsuzuku2018lipschitz}.
Parseval networks~\citep{cisse2017parseval} ensure weight matrices are 1- Lipschitz by enforcing a stronger constraint, orthogonality.~\citet{bartlett2018representing} show that
any bi-Lipschitz function can be wrriten as a compositions of residual layers~\citep{resnets}.

Instead of restricting the learned function on the entire space, another approach of targeting smoothness constraints is to regularize the norm of the gradients with respect to inputs of the network $\norm{\nabla_{\vx}f_{\vtheta}(\vx)}$, at different \textit{regions of the space}~\citep{sokolic2017robust,gulrajani2017improved,fedus2017many,arbel2018gradient,kodali2017convergence}.
 This is often enforced by adding a gradient penalty to the loss function $\mathcal{L(\vtheta)}$:
\begin{align}
  \mathcal{L(\vtheta)} + \lambda \mathbb{E}_{p_{reg(\vx)}} \left(\norm{\nabla_{\vx}f_{\vtheta}(\vx)}_2^2 - K^2\right)^2
\label{eq:grad_penalties}
\end{align}

where $\lambda$ is a regularization coefficient,
$p_{reg(\vx)}$ is the distribution at which the regularization is applied,
 which can either be the data distribution~\citep{sokolic2017robust,arbel2018gradient} or around it~\citep{kodali2017convergence,fedus2017many},
 or, in the case of generative models, at linear interpolations between data and model samples~\citep{gulrajani2017improved}.
 Gradient penalties encourage the function to be smooth around the support of $p_{reg(\vx)}$ either by encouraging
 Lipschitz continuity ($K \ne 0$) or by discouraging drastic changes of the function as the input changes ($K=0$).

Smoothness for classification tasks is defined by \citet{lassance2018laplacian} as preserving features similarities within the same class as we advance through the layers of the network.
The penalty used is $\sum_{l=1}^{L} \sum_{c=1}^{C}|\sigma^l(s_c) - \sigma^{l+1}(s_c)|$, where $\sigma^l(s_c)$ is the signal of features belonging to class $c$ computed using the Laplacian of layer $l$.
The Laplacian of a layer is defined by constructing a weighted symmetric adjacency matrix of the graph induced by the pairwise most similar layer features in the dataset. This type of task dependent approach to smoothness is promising, as we will discuss later.

\section{The benefits of smooth function approximators}
\label{sec:benefits}

\textbf{Generalization}. Learning models that generalize beyond training data is the goal of machine learning.
The common wisdom is that models with small complexity generalize better~\citep{vapnik2013nature}.
Despite this, we have seen that deep, overparametrized neural networks tend to generalize better~\citep{novak2018sensitivity} and that for Bayesian methods, Occam's razor does not apply to the number of parameters used, but to the complexity of the function~\citep{rasmussen2001occam}.
A way to reconcile these claims is to incorporate smoothness into definitions of model complexity and to show that smooth, overparametrized neural networks generalize better than their less smooth counterparts.
Methods that encourage smoothness such as weight decay, dropout and early stopping have been long shown to aid generalization~\citep{bartlett1997valid,golowich2018size,train_faster_generalize_better,deep_double_descent,dropout}.
Data augmentation
 has been shown to increase robustness to random noise or to modality specific transformations such as image cropping and rotations~\citep{krizhevsky2012imagenet,simonyan2014very,data_augumentation_generative_modeling}.
~\citet{sokolic2017robust} show that the generalization error of a network with linear, softmax and pooling layers is bounded by the classification margin in input space.
Since classifiers are trained to increase classification margins in output space, smoothing by bounding the spectral norm of the model's Jacobian increases generalization performance; this leads to empirical gains on standard image classification tasks.

Generalization has been recently reexamined under the light of double descent~\citep{deep_double_descent,double_descent},
a phenomenon named after the shape of the generalization error
plotted against the size of a deep neural network: as the size of the network increases the generalization error decreases (first descent), then increases, after which it decreases again (second descent).
We postulate there is a deep connection between double descent and smoothness:
in the first descent, the generalization error is decreasing as the model is given extra capacity to
capture the decision surface; the increase happens when the model has enough capacity to fit the training data, but it cannot do so and retain smoothness; the second descent occurs as the capacity increases and smoothness can be retained.
 This view of double descent is supported by empirical evidence which shows that its effect
is most pronounced on clean label datasets and when early stopping and other regularization techniques are not used~\citep{deep_double_descent}. We later show that smoothness constraints heavily interact with optimization which further suggests that empirical investigations into the impact of smoothness on the observation of double descent are needed.

\textbf{Reliable uncertainty estimates}. Neural networks trained to minimize classification losses provide notoriously unreliable uncertainty estimates;
an issue which gets compounded when the networks are faced with out of distribution data.
However, one can still leverage the power of neural networks to obtain reliable uncertainty estimates, by combining smooth neural feature learners with non-softmax decision surfaces
~\citep{van2020simple,balaji_uncertanty_smoothness}.
The choice of smoothness regularization or classifier can vary,
from using
 gradient penalties on the neural features with a Radial Basis Function classifier~\citep{van2020simple},
to using Spectral Normalization on neural features
and a Gaussian Process classifier~\citep{balaji_uncertanty_smoothness}.
These methods
are competitive with standard techniques used for out of distribution detection~\citep{lakshminarayanan2017simple}
 on both vision and language understanding tasks.
The importance of smoothness regularizing neural features indicates that having a smooth decision surface such as a Gaussian Process is not sufficient to compensate for sharp feature functions when learning models for uncertainty estimation.

\textbf{Robustness to adversarial attacks}. Adversarial robustness has become an active area of research in recent years~\citep{szegedy2013intriguing,goodfellow2014explaining,papernot2016limitations,chakraborty2018adversarial}.
Early works have observed that the existence of adversarial examples is related to the magnitude of the
gradient of the hidden network activation with respect to its input, and suggested that constraining
the Lipschitz constant of individual layers can make networks more robust to attacks~\citep{szegedy2013intriguing}.
However, initial approaches to combating adversarial attacks focused on data augmentation methods~\citep{kurakin2016adversarial,goodfellow2014explaining,moosavi2017universal,madry2017towards},
and only more recently smoothness constraints have come into focus~\citep{cisse2017parseval,novak2018sensitivity,sokolic2017robust,lassance2018laplacian}.
We can see the connection between smoothness and robustness by looking at the desired robustness properties of classifiers, which aim to ensure that inputs in the same $\epsilon$-ball result in the same function output:
\begin{align}
 \norm{x - x'} \le \epsilon \implies \arg \max f(\vx) = \arg \max f(\vx')
 \label{eq:adv_examples}
\end{align}
The aim of adversarial defenses and robustness techniques is
to have $\epsilon$ be as large as possible without affecting classification accuracy.
Robustness against adversarial examples has been shown to correlate with generalization~\citep{gilmer2018adversarial}, and with the sensitivity of the network output with respect to the input as measured by the Frobenius norm of the Jacobian of the learned function~\citep{novak2018sensitivity,sokolic2017robust}.
~\citet{lassance2018laplacian} show that robustness to adversarial examples is enhanced when the function approximator is smooth as defined by the Laplacian smoothness signal discussed in Section~\ref{sec:smoothness_techniques}.
~\citet{tsuzuku2018lipschitz} show that  Equation~\ref{eq:adv_examples}
holds when the $l_2$ norm is used if $\epsilon$ is smaller than the ratio of the classification margin and the
Lipschitz constant of the network times a constant, and thus they increase robustness by ensuring the margin is larger than the Lipschitz constant.

\textbf{Improved generative modeling performance}.
Smoothness constraints through gradient penalties or spectral normalization have become a
recipe for obtaining state of the art generative models.
In generative adversarial networks (GANs)~\citep{goodfellow2014generative}, smoothness constraints on
the discriminator and the generator have played a big part in scaling up training on large,
diverse image datasets at high resolution~\citep{biggan} and a combination of smoothness
constraints has been shown to be a requirement to get GANs to work on discrete data such as text~\citep{scratch_gan}.
The latest variational autoencoders~\citep{kingma2013auto,rezende2014stochastic} incorporate spectral regularization to boost performance and stability~\citep{vahdat2020nvae}.
Explicit likelihood tractable models like normalizing flows~\citep{normalizing_flows} benefit from smoothness constraints through powerful invertible layers built using residual connections
$g(\vx) = \vx + f(\vx)$ where $f$ is Lipschitz
 ~\citep{invertible_residual_networks}.

\textbf{More informative critics}. Critics, learned approximators to intractable decision functions, have become a fruitful endeavor in generative modeling, representation learning and reinforcement learning.

\begin{figure}[t]
  \centering
  \begin{subfigure}[b]{0.44\textwidth}
  \includegraphics[width=\textwidth]{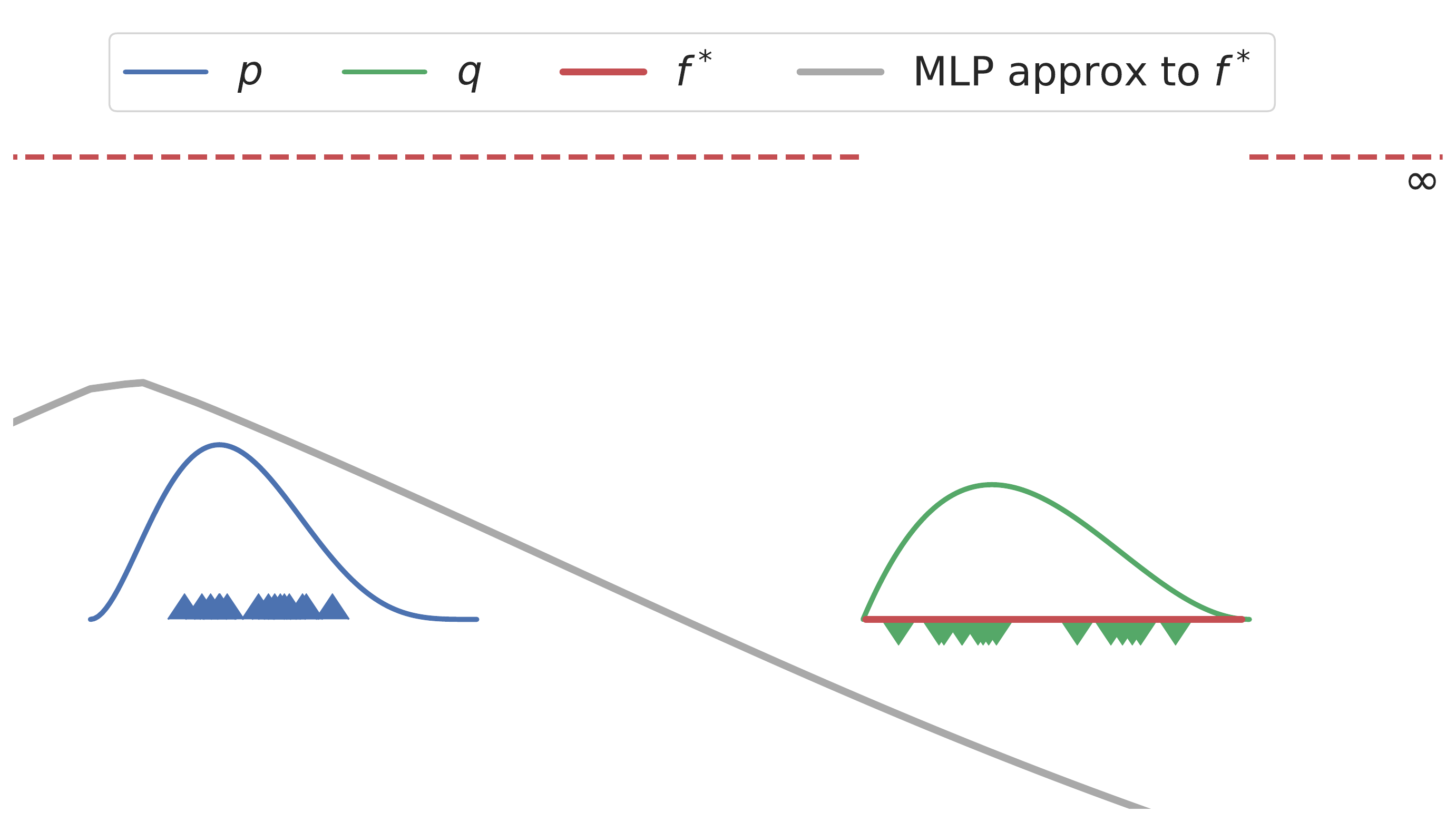}
  \caption{KL divergence optimal critic $f^* = \frac{p}{q}$ \\and smooth critic estimate. }
  \label{fig:kale_gan}
  \end{subfigure}
  \begin{subfigure}[b]{0.44\textwidth}
  \includegraphics[width=\textwidth]{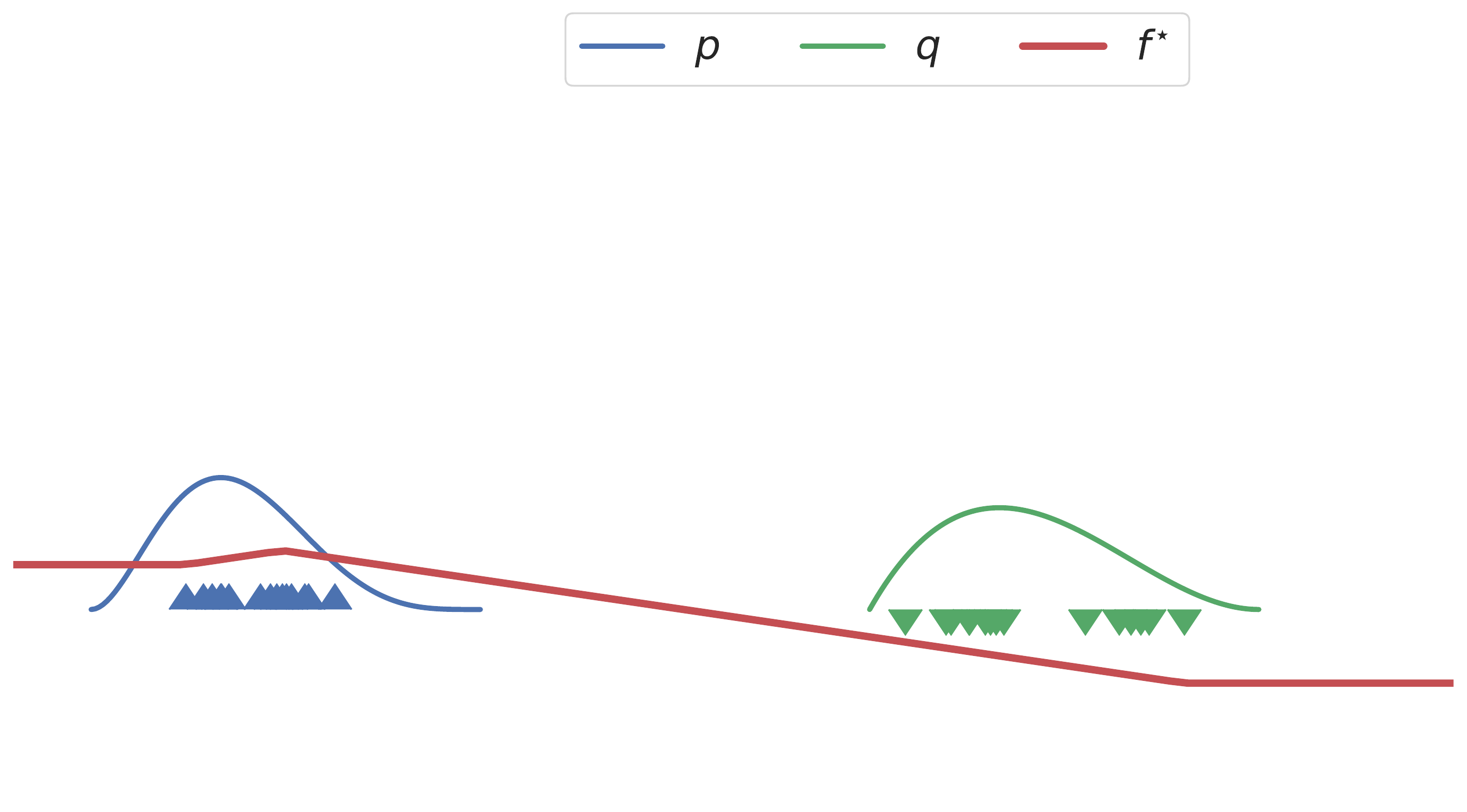}
  \caption{Optimal Wasserstein critic is smooth.\newline}
  \label{fig:optimal_critic_wasserstein}
  \end{subfigure}
  \caption{The importance of critic smoothness when estimating divergences and distances. Left: When the two distributions do not have overlapping support, the KL divergence provides no learning signal, while a smooth approximation via a learned critic does. Right: The optimal Wasserstein critic has a smoothness Lipschitz constraint in its definition.}
\end{figure}

Critics are used in generative modeling to approximate divergences and distances between the learned model and the true unknown
data distribution, and have been mainly popularised by GANs.
A critic in a function class $\mathcal{F}$ can be used to approximate the KL divergence by minimizing the bound~\citep{nguyen2010estimating,f_gan}:
\begin{align*}
\mathrm{KL}(p||q) = \mathbb{E}_{p(\vx)} \log\frac{p(\vx)}{q(\vx)} = \sup_{f} \mathbb{E}_{p(\vx)} f(\vx) - \mathbb{E}_{q(\vx)} e^{f(\vx)-1} \ge \sup_{f \in \mathcal{F}} \mathbb{E}_{p(\vx)} f(\vx) - \mathbb{E}_{q(\vx)} e^{f(\vx)-1}
\end{align*}
While due to the density ratio $p(\vx)/q(\vx)$ in its definition, the KL divergence provides no learning signal when the model and data distributions do not have overlapping support, choosing $\mathcal{F}$ to be a family of smooth functions results in a bound on the KL which provides useful gradients and can be used to train a model~\citep{fedus2017many,arbel2020kale}.
 We show an illustrative example in Figure~\ref{fig:kale_gan}: the true decision surface jumps from zero to infinity,
 while the approximation provided by the MLP is smooth.
Similarly, training the critic more and making it better at estimating the true decision surface but less smooth can hurt training~\citep{schafer2019implicit}. It's not surprising that imposing smoothness constraints on critics has become part of many flavours of GANs~\citep{arjovsky2017wasserstein,gulrajani2017improved,fedus2017many,biggan,arbel2018gradient,lipschitz_gan,yoshida2017spectral}.

The same conclusions have been reached in unsupervised representation learning, where
parametric critics are trained to approximate another intractable quantity, the mutual information, using
 the Donsker–Varadhan or similar bounds~\citep{belghazi2018mine,cpc}.
An extensive study on representation learning techniques based on mutual information showed that tighter bounds do not lead to better representations~\citep{on_mi_for_rep_learning}. Instead, the success of these methods is attributed to the inductive biases of the critics employed to approximate the mutual information.
In reinforcement learning, neural function approximators or ``critics'' approximate
state-value functions or action-state value functions and are then used to train a policy to maximize the expected reward. Directly learning a neural network parametric estimator of the action value \textit{gradients}
 - the gradients of the action value with respect to the action - results in more accurate gradients (Figure 3 in \citep{d2020learn}),
 but also makes gradients smoother.
 This provides an essential exploration prior in continuous control,
 where similar actions likely result in the same reward and observing the same action twice is
 unlikely due to size of the action space;
 encouraging the policy network to extrapolate from the closest seen action
improves performance over both model free and model based continuous control approaches~\citep{d2020learn}.

\textbf{Distributional distances}. Including smoothness constraints in the definition of distributional distances by using optimal transport has seen a uptake in machine learning applications in recent years,
from generative modeling~\citep{arjovsky2017wasserstein,gulrajani2017improved,wasserstein_autoencoders,ostrovski2018autoregressive}
to reinforcement learning~\citep{bellemare2017distributional,dabney2018implicit}, neural ODEs~\citep{finlay2020train}
 and fairness~\citep{chiappa2020general,jiang2020wasserstein}.
Optimal transport is connected to Lipschitz smoothness as the Wasserstein distance can be computed via the Kantorovich-Rubinstein duality~\citep{villani2008optimal}:
\begin{align}
W_1(p(\vx), q(\vx)) = \sup_{f: \norm{f}_{\text{Lip}} \le 1} \mathbb{E}_{p(\vx)} f(\vx) - \mathbb{E}_{q(\vx)} f(\vx)
\label{eq:wasserstein}
\end{align}

The Wasserstein distance is finding the critic that can separate the two distributions in expectation,
but constraints that critic to be Lipschitz in order to avoid pathological solutions.
The importance of the Lipschitz constraint on the critic can be seen in Figure~\ref{fig:optimal_critic_wasserstein}:
unlike the KL divergence,
the optimal Wasserstein critic is well defined when the two distributions do not have overlapping support, and does not require an approximation to provide useful learning signal for a generative model.

\section{Consequences of poor smoothness assumptions}

\textbf{Weak models}. Needlessly limiting the capacity of our models by enforcing smoothness constraints is a significant danger:
 a constant function is very smooth, but not very useful.
Beyond trivial examples, \citet{excessive_invariance_adv_vulnerability} show that one of the reasons  neural networks are vulnerable to adversarial perturbations is invariance to task relevant changes - too much smoothness with respect to the wrong metric. A neural network can be ``too Lipschitz'':
methods aimed at increasing robustness to adversarial examples do indeed decrease the Lipschitz constant of a classifier,
but once the Lipschitz constant becomes too low, accuracy drops significantly~\citep{fazlyab2019efficient}.

There are two main avenues for being too restrictive in the specification of smoothness constraints,
depending on \textit{where} and \textit{how} smoothness is encouraged.
  Smoothness constraints can be imposed on the entire input space or
only in certain pockets, often around the data distribution.
Methods which impose constraints on the entire space throw away useful information about the input distribution and restrict the learned function needlessly by forcing it to be smooth in areas of the space where there is no data. This is especially problematic when the input lies on a small manifold in a large dimensional space, such as in the case of natural images, which are a tiny fraction of the space of all possible images.
Model capacity can also be needlessly restrained by imposing strong constraints on the individual components of the model, often the network layers, instead of allowing the network to allocate capacity as needed.

\begin{figure}[t]
  \centering
  \begin{subfigure}[b]{0.3\textwidth}
  \includegraphics[width=\textwidth]{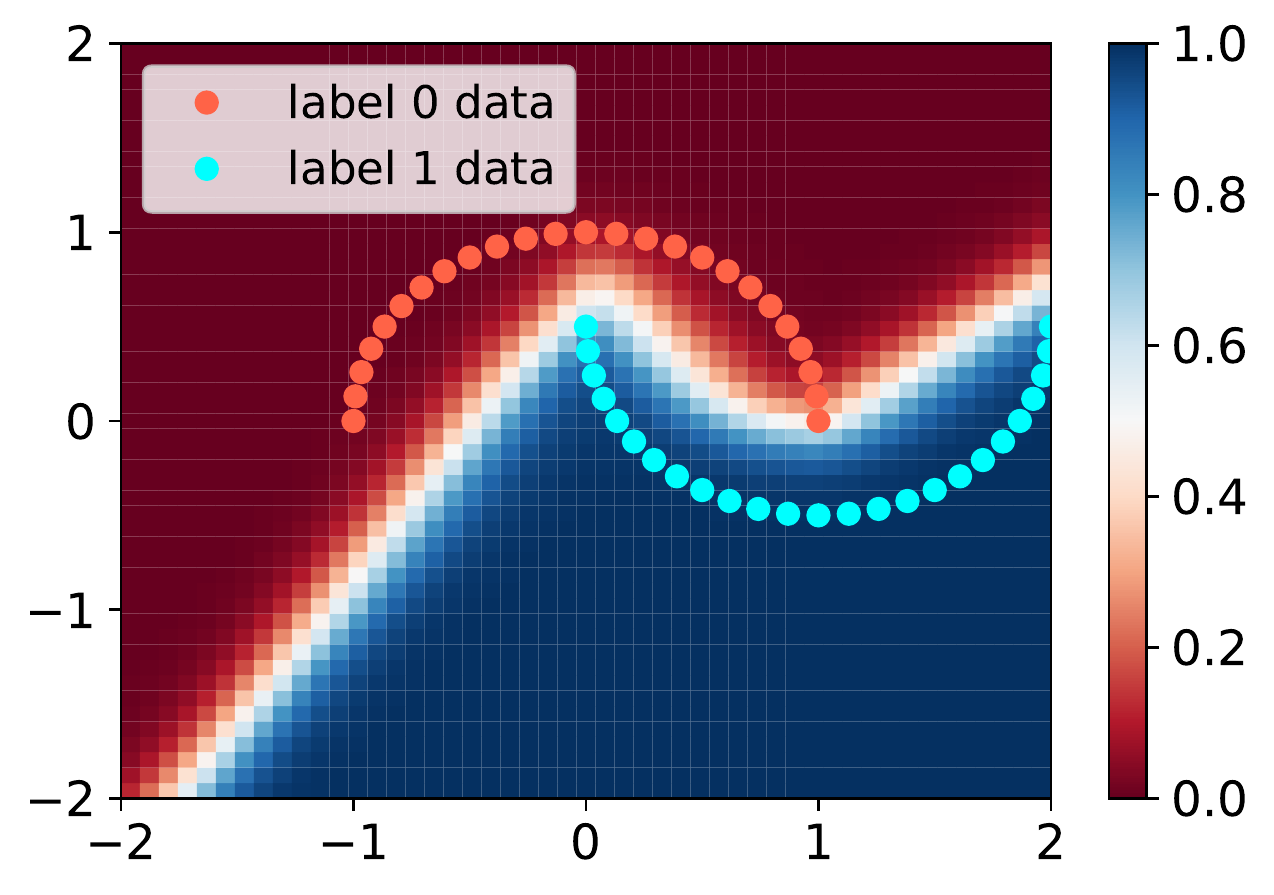}
  \caption{2 layer MLP.\newline}
  \label{fig:two_moons_decision_surface_shallow_mlp}
  \end{subfigure}
  \begin{subfigure}[b]{0.32\textwidth}
  \includegraphics[width=\textwidth]{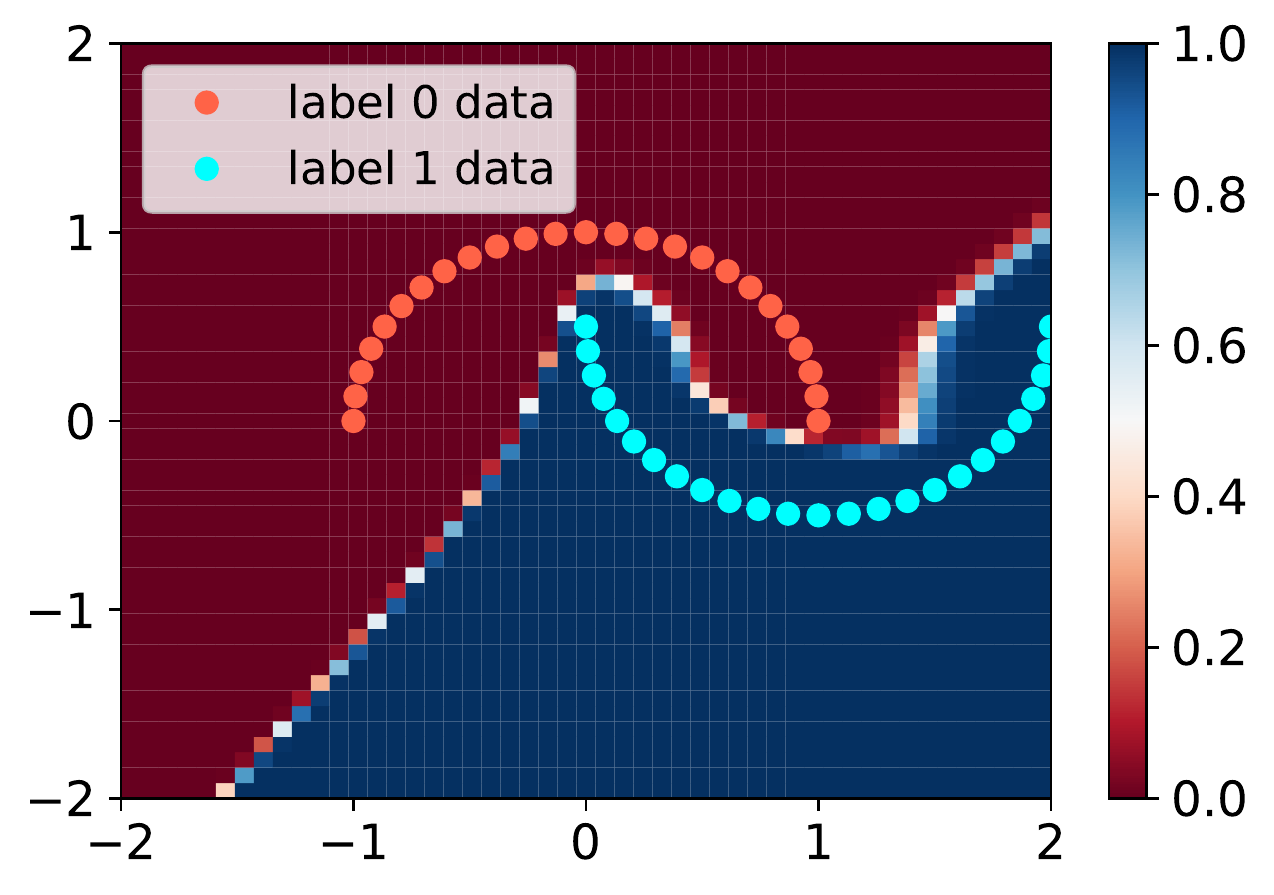}
  \caption{4 layer MLP.\newline}
  \label{fig:two_moons_decision_surface_deep_mlp}
  \end{subfigure}
  \begin{subfigure}[b]{0.32\textwidth}
  \includegraphics[width=\textwidth]{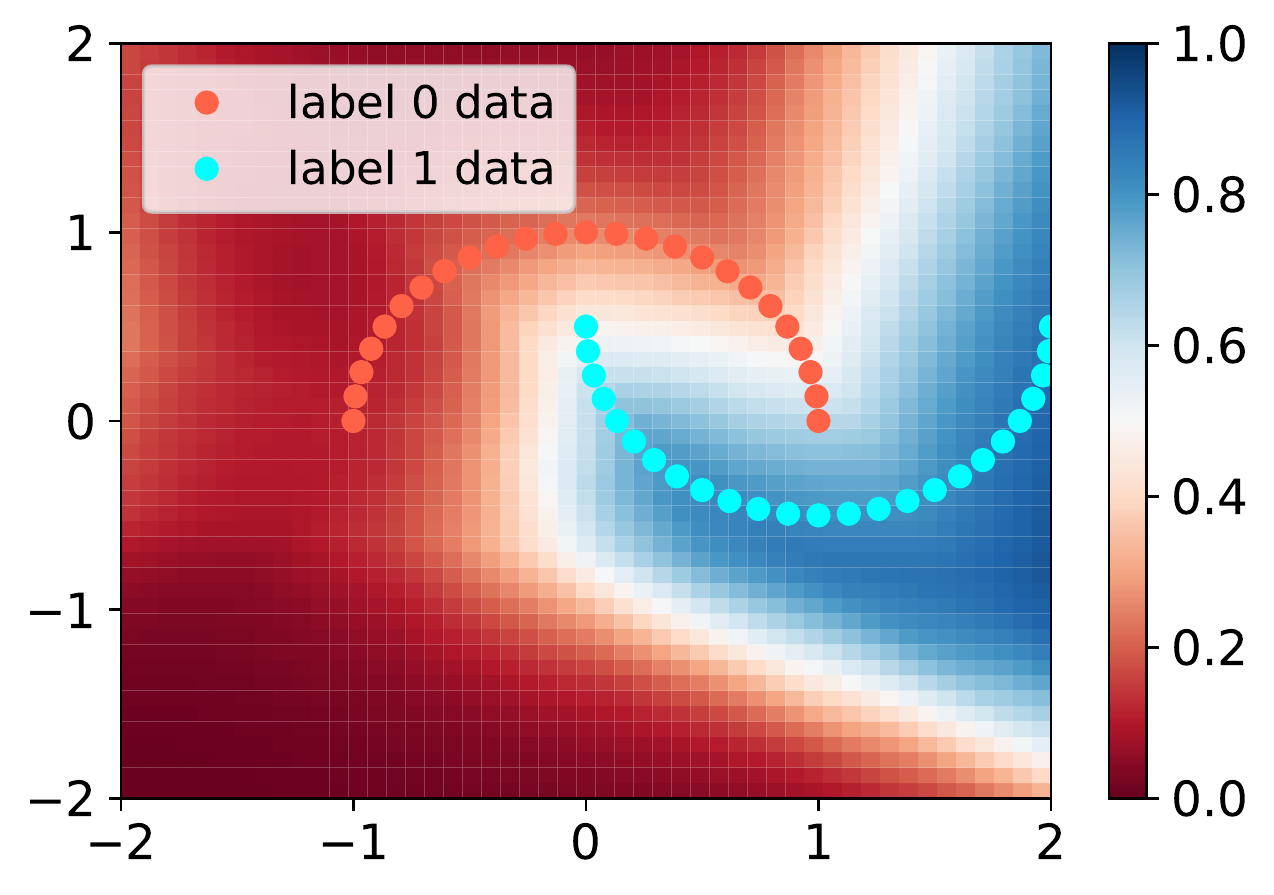}
  \caption{Gradient penalty at data; $K=1$.}
    \label{fig:two_moons_decision_surface_grad_penalty}
  \end{subfigure}\\
  \begin{subfigure}[b]{0.32\textwidth}
  \includegraphics[width=\textwidth]{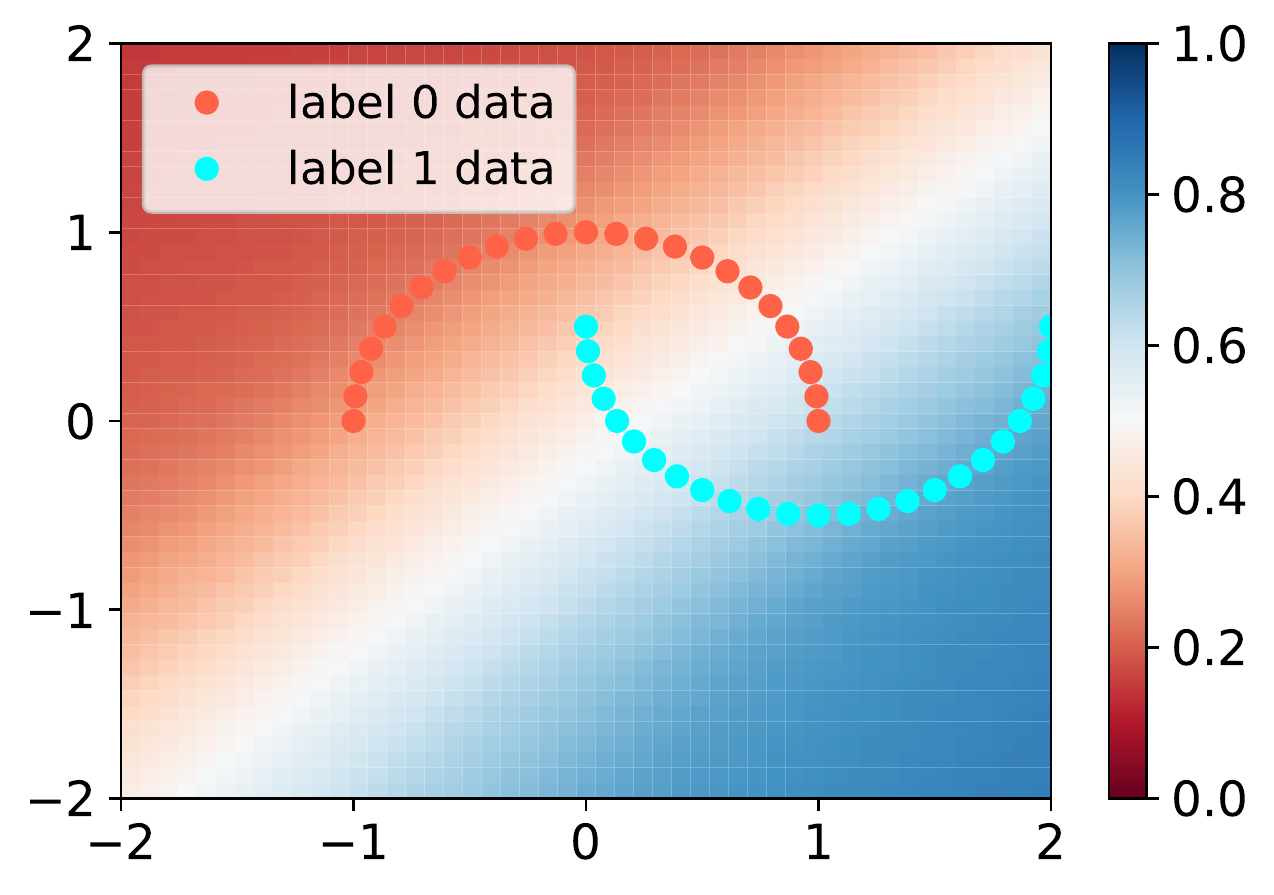}
  \caption{Spectral norm; \\$K=1$.}
  \label{fig:two_moons_decision_surface_sn}
  \end{subfigure}
  \begin{subfigure}[b]{0.32\textwidth}
  \includegraphics[width=\textwidth]{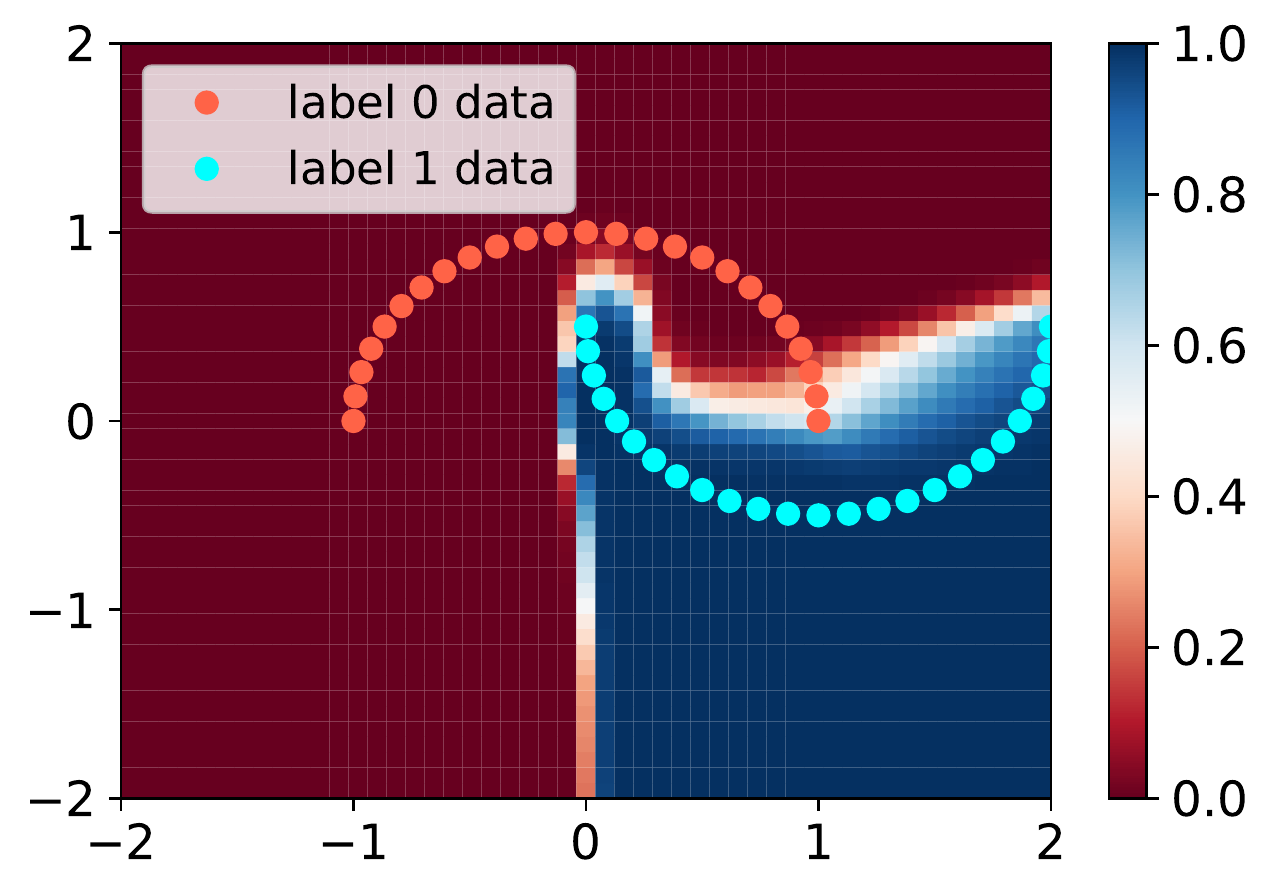}
  \caption{Spectral norm; \\$K=10$.}
  \label{fig:two_moons_decision_surface_sn_10}
  \end{subfigure}
  \begin{subfigure}[b]{0.32\textwidth}
  \includegraphics[width=\textwidth]{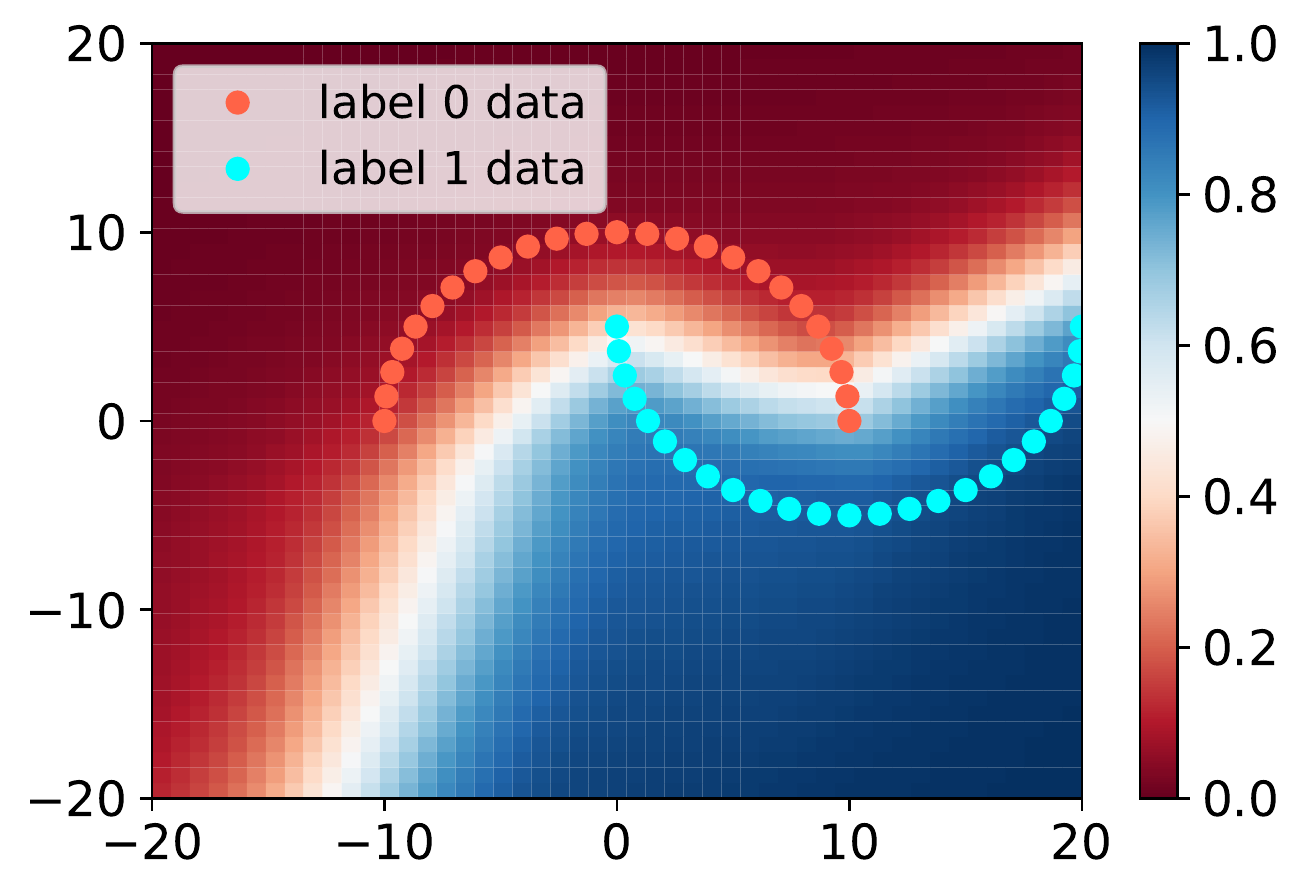}
  \caption{Spectral norm; \\scaling the data by 10.}
    \label{fig:two_moons_decision_surface_sn_data_scaling}
  \end{subfigure}
  \caption{Decision surfaces on \textit{two moons} under different regularization methods.\\ Unless otherwise specified the model architecture is a 4 layer MLP.}
  \label{fig:two_moons_decision_surface}
\end{figure}

\begin{figure}[t]
  \centering
  \begin{subfigure}[b]{0.32\textwidth}
  \includegraphics[width=\textwidth]{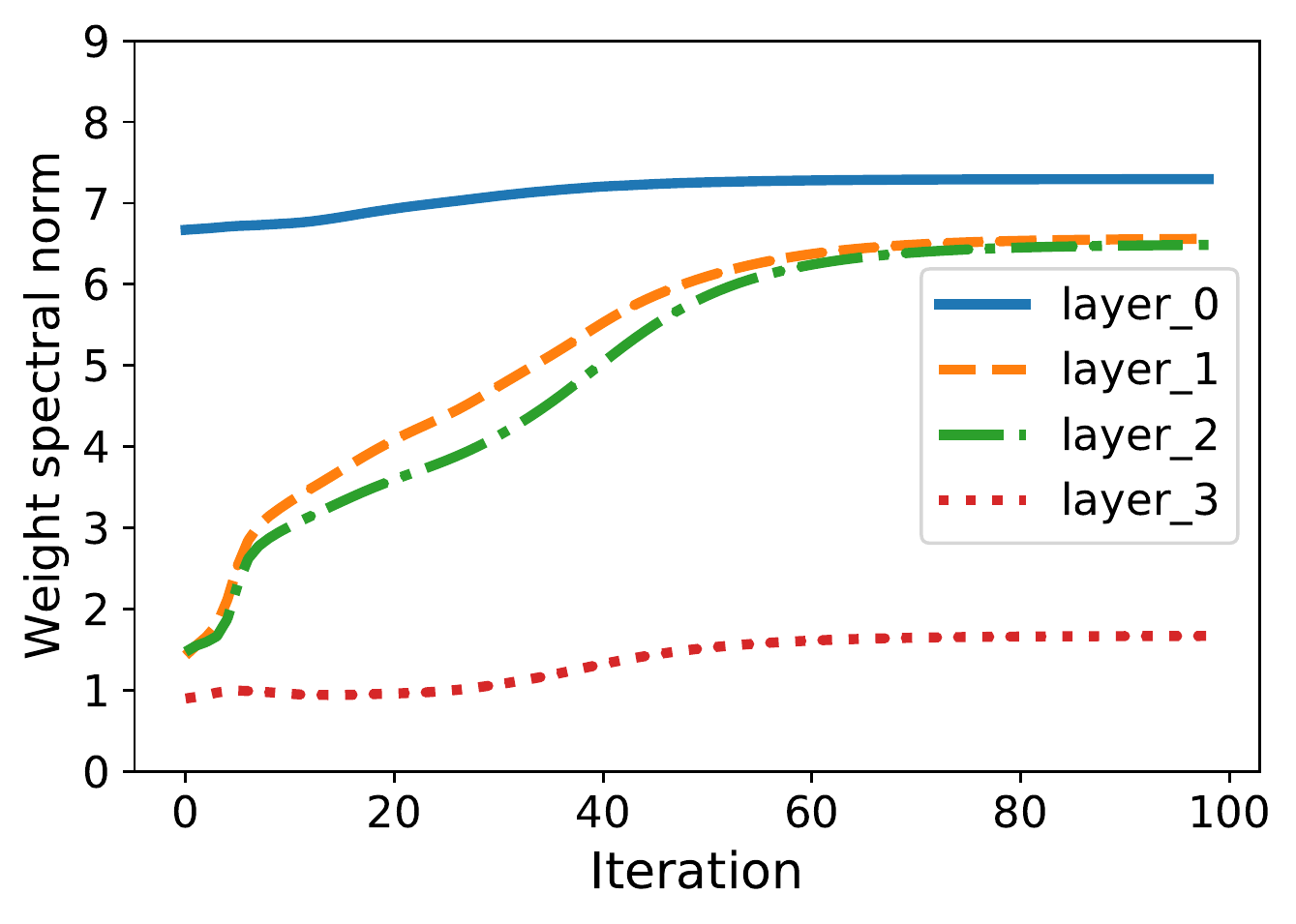}
  \caption{Unregularized.}
  \end{subfigure}
  \begin{subfigure}[b]{0.32\textwidth}
  \includegraphics[width=\textwidth]{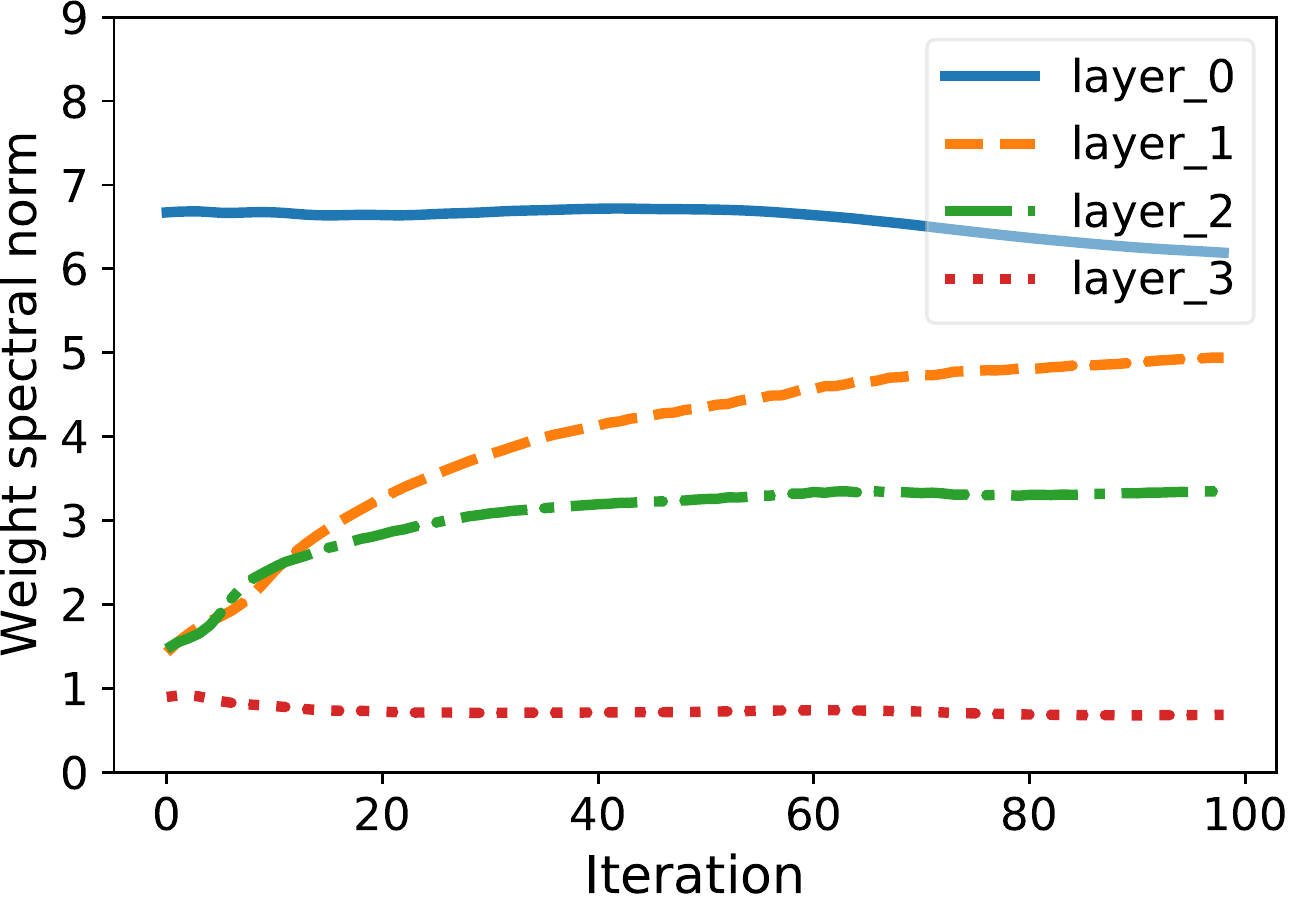}
  \caption{Gradient penalty. $K=1$.}
  \end{subfigure}
  \begin{subfigure}[b]{0.32\textwidth}
  \includegraphics[width=\textwidth]{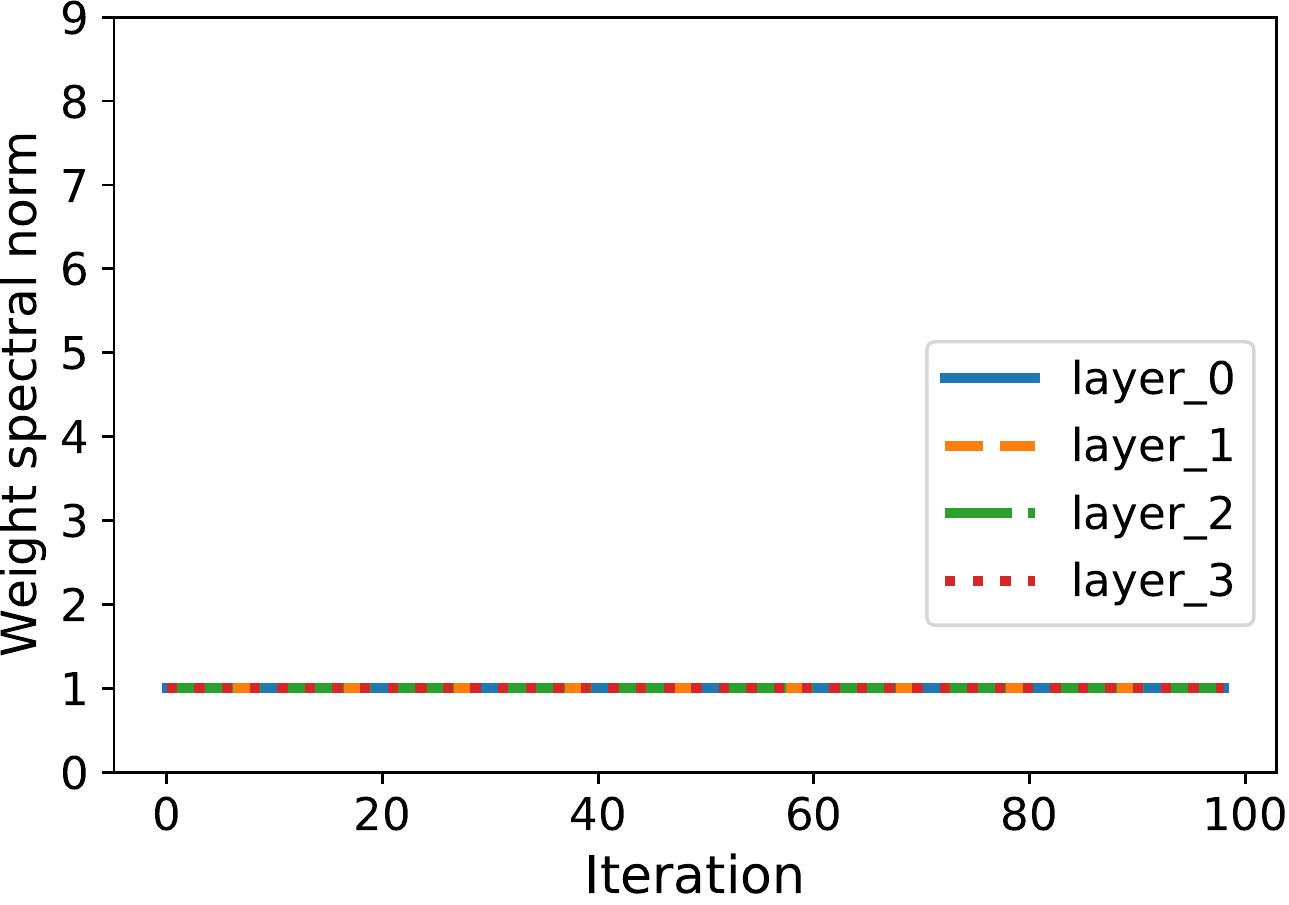}
  \caption{Spectral Normalization. $K=1$.}
  \end{subfigure}

  \caption{Lipschitz constant of each layer of an MLP trained on the two moons dataset. \\The decision surfaces for the same models can be seen in Figure~\ref{fig:two_moons_decision_surface}. Smaller means smoother.}
  \label{fig:two_moons_layers_constant}
\end{figure}

\begin{figure}[t]
  \centering
  \begin{subfigure}[b]{0.32\textwidth}
  \includegraphics[width=\textwidth]{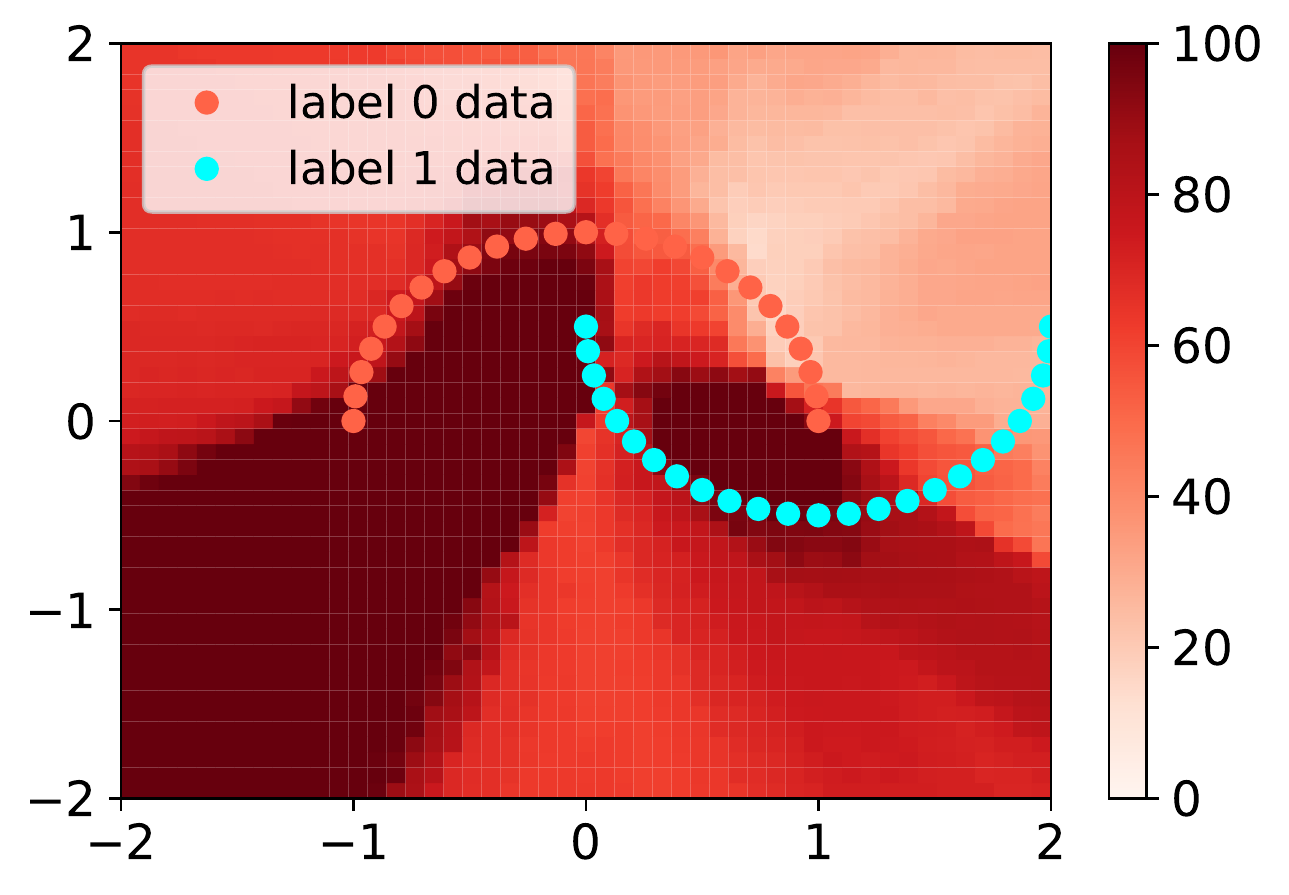}
  \caption{Unregularized.}
  \end{subfigure}
  \begin{subfigure}[b]{0.32\textwidth}
  \includegraphics[width=\textwidth]{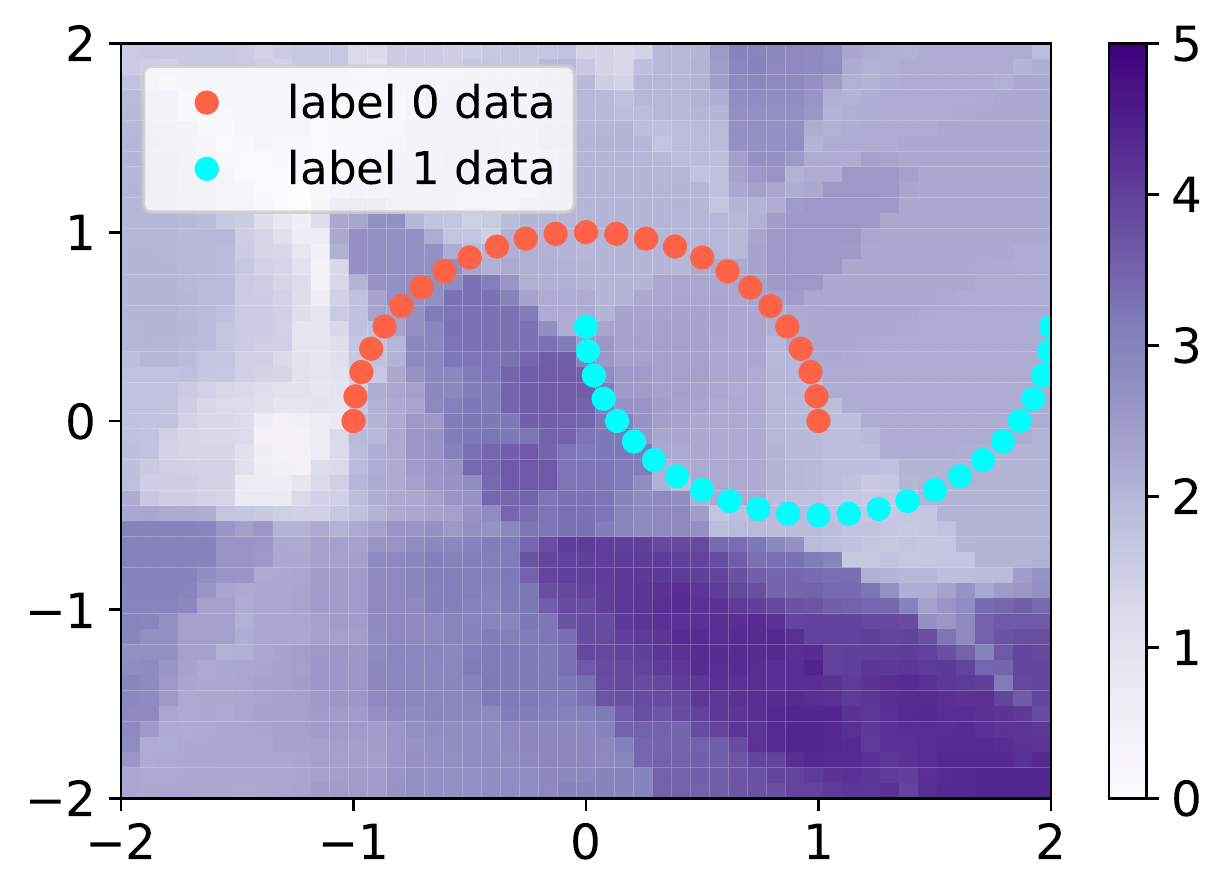}
  \caption{Gradient penalty. $K=1$.}
  \end{subfigure}
  \begin{subfigure}[b]{0.32\textwidth}
  \includegraphics[width=\textwidth]{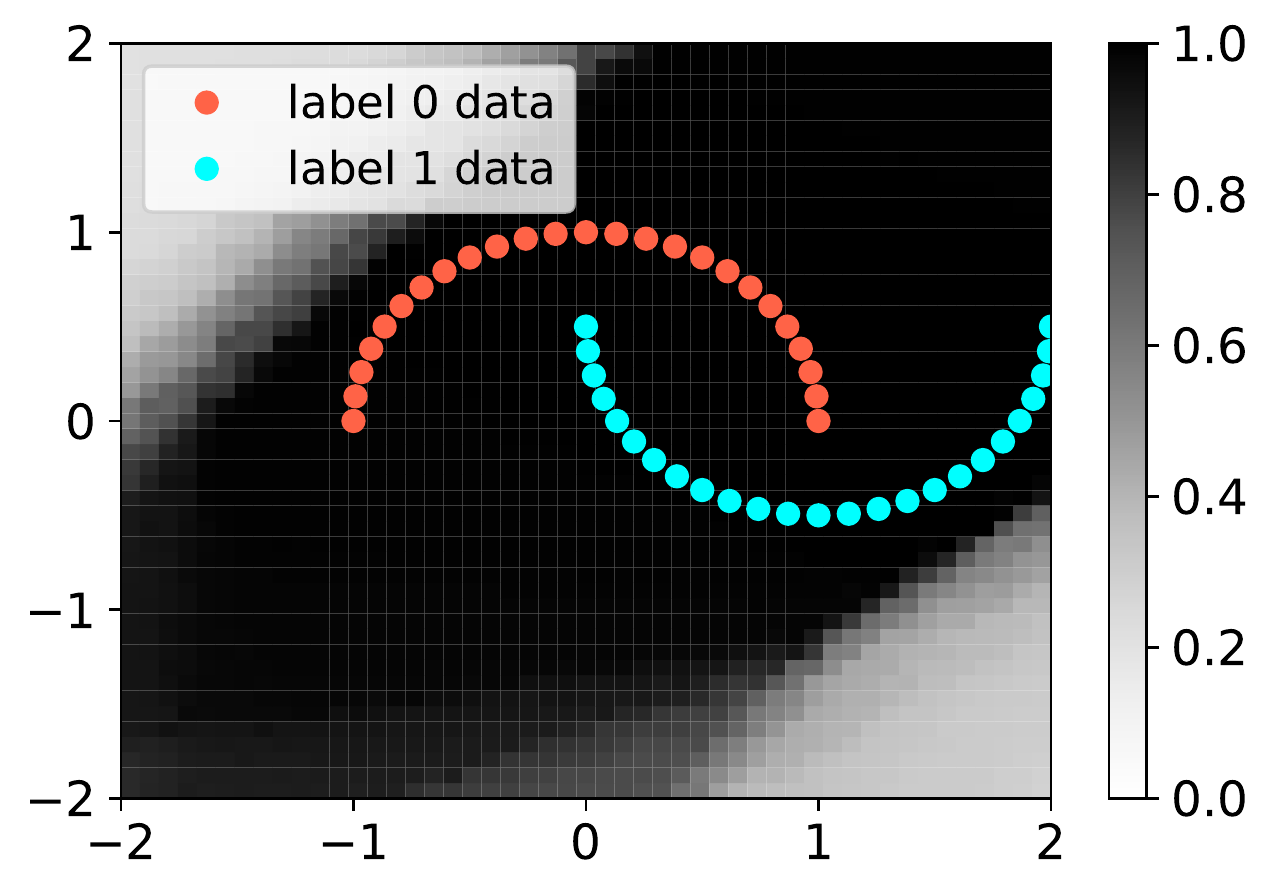}
  \caption{Spectral Normalization. $K=1$.}
  \end{subfigure}
  \captionsetup{justification=justified}
  \caption{The effect of regularization on \textit{local} smoothness. We plot the local Lipschitz constants obtained this using an exhaustive grid search in \textit{local neighborhoods}, instead of loose bounds. We use different colors to emphasize the different scale of the constants for the different methods.}
  \label{fig:two_moons_layers_local_constant}
\end{figure}

We can exemplify the importance of where and how constraints are imposed with an example, by contrasting gradient penalties - end to end regularization applied around the training data - and Spectral Normalization - layerwise regularization applied to the entire space. Figure~\ref{fig:two_moons_decision_surface_sn} shows that using Spectral Normalization to restrict the Lipschitz constant of an MLP to be 1 decreases the capacity of the network and severely affects accuracy compared to the
 baseline MLP - Figure~\ref{fig:two_moons_decision_surface_deep_mlp} - or the MLP regularized using gradient penalties -
 Figure~\ref{fig:two_moons_decision_surface_grad_penalty}.
 Further insight comes from Figure~\ref{fig:two_moons_layers_constant}, which shows that the gradient penalty only enforces a
weak constraint on the model and does not heavily restrict the spectral norms of individual layers; this is in stark contrast with Spectral Normalization which by construction ensures each network layer has spectral norm equal to 1. To show the effect of data dependent regularization on \textit{local smoothness} we plot the Lipschitz constants of the model at neighborhoods spanning the entire space in Figure~\ref{fig:two_moons_layers_local_constant}. Each Lipschitz constant is
computed using an exhaustive grid search inside each local neighborhood rather than a bound -
details are provided in Appendix~\ref{app:exp}.
As expected, gradient penalties impose stronger constraints around the training data, while Spectral Normalization has a strong effect on the smoothness around points in the entire space.
This simple example suggests that the search for better smoothness priors needs
to investigate \textit{where} we want functions to be smooth
and reexamine \textit{how} smoothness constraints should account for the compositional aspect of neural networks,
otherwise we run the risk of learning trivially smooth functions.

\textbf{Overlooked interactions with  optimization}.
We show that viewing smoothness only through the lens of the model is misleading, as smoothness constraints have a strong effect on optimization.
The interaction between smoothness and optimization has been mainly observed when training generative models;
 encouraging the smoothness of the encoder through spectral regularization increased the stability of hierarchical VAEs and led variational inference models to the state of the art of explicit likelihood non autoregressive models~\citep{vahdat2020nvae},
 while smoothness regularization of the critic (or discriminator)
has been established as an indispensable stabilizer of GAN training, independently of the training criteria used~\citep{yoshida2017spectral,arbel2018gradient,fedus2017many,biggan,kurach2019large}.

\begin{figure}[t]
  \centering
     \begin{subfigure}[b]{0.33\textwidth}
     \renewcommand\thesubfigure{a}
      \includegraphics[width=\textwidth]{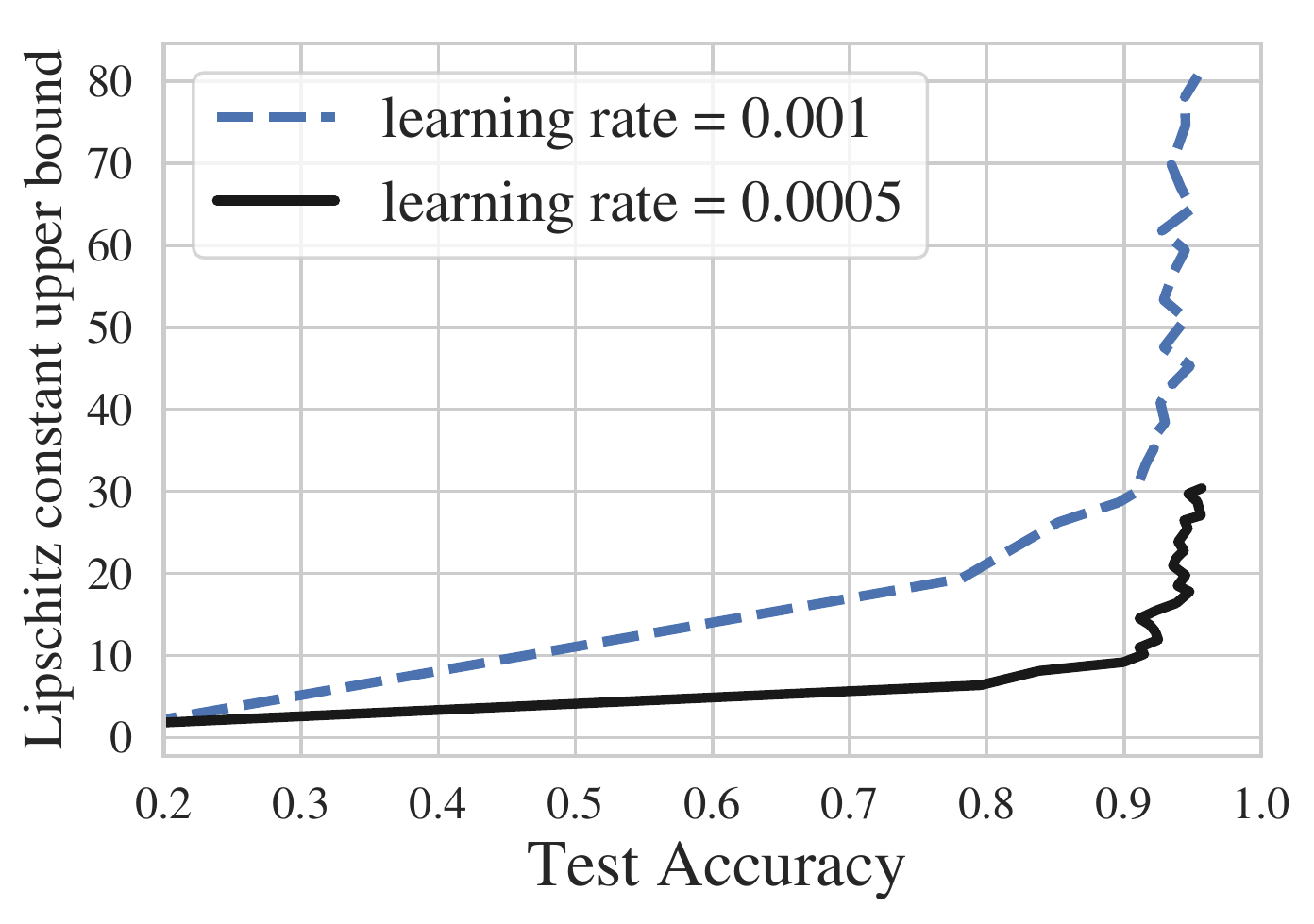}
        \caption{MNIST classification.}
     \label{fig:mnist_classification}
  \end{subfigure}
  \begin{subfigure}[b]{0.65\textwidth}
    \renewcommand\thesubfigure{b}
    \begin{subfigure}[b]{0.48\textwidth}
    \includegraphics[width=\textwidth]{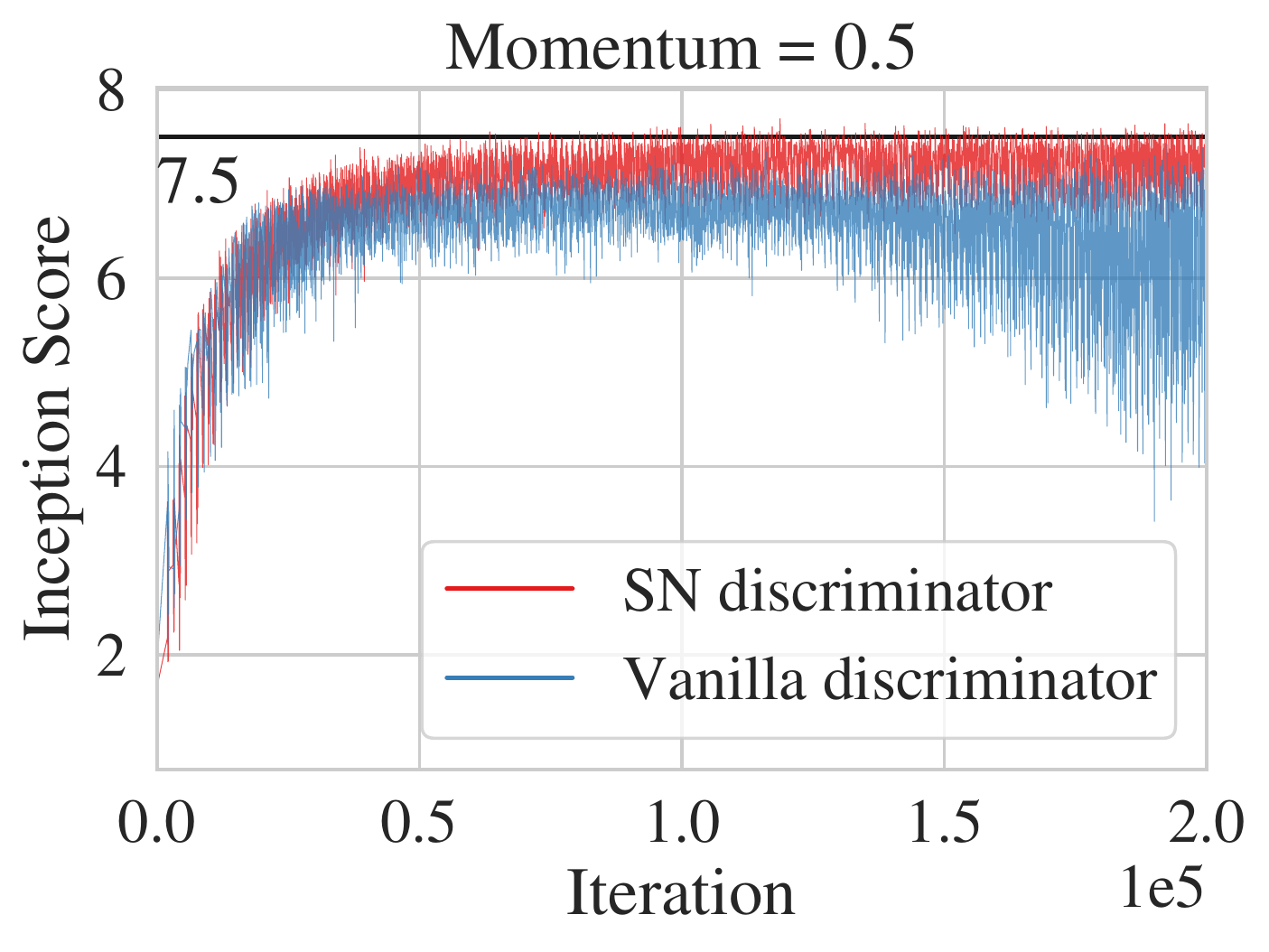}
    \end{subfigure}%
    \begin{subfigure}[b]{0.48\textwidth}
      \renewcommand\thesubfigure{a2}
      \includegraphics[width=\textwidth]{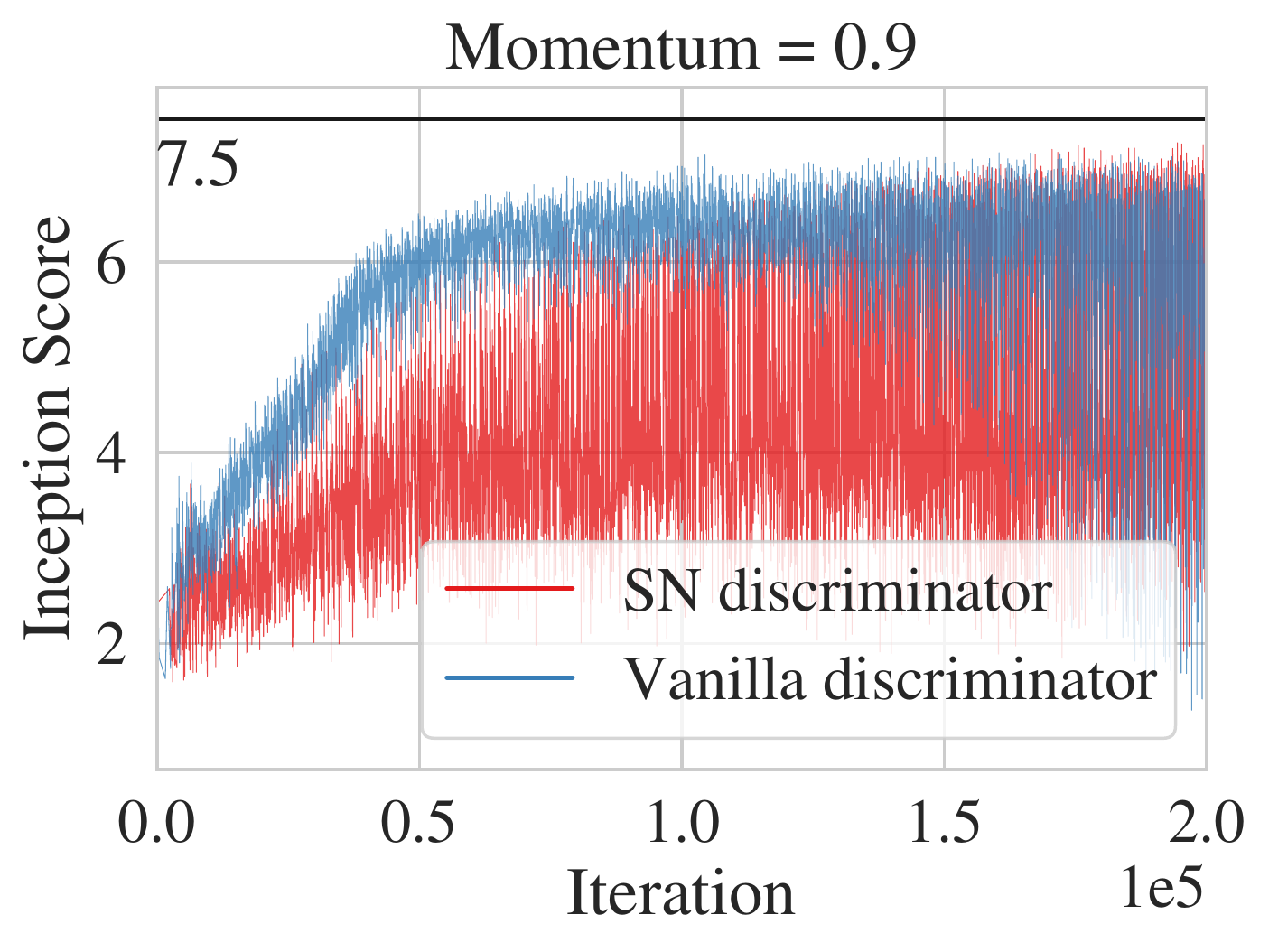}
    \end{subfigure}%
  \caption{Spectral Normalization requires low momentum in GAN training.}
  \label{fig:sn_mom}
  \end{subfigure}%
  \captionsetup{justification=justified}
  \caption{Smoothness interacts with optimization. Left: MNIST classifiers trained with different learning rates exhibit different smoothness behaviour throughout training, even at the same test accuracy. Lower means smoother. Right: The effects of momentum on Spectral Normalization applied to the GAN discriminator on CIFAR-10. Higher is better.}

\end{figure}

Some smoothness regularization techniques affect optimization by
changing the loss function (gradient penalties, spectral regularization) or the optimization regime directly (early stopping, projection methods). Even if they don't explicitly change the loss function or optimization regime,
smoothness constraints affect the path the model takes to reach convergence.
We use a simple example to show why smoothness regularization interacts with optimization in Figure~\ref{fig:mnist_classification}. We use different learning rates to train two unregularized MLP classifiers on MNIST~\citep{lecun1999object}  and observe that the learning rate used affects its smoothness throughout training, without changing testing accuracy. This shows that
imposing similar smoothness constraints on two models which share the same architecture but are trained with  different learning rates would lead to very different strengths of regularization and drastically change the trajectory of optimization.

Beyond learning rates, smoothness constraints also interact with momentum.
In the GAN setting,~\citet{gulrajani2017improved} observed that weight clipping in the Wasserstein GAN critic requires low to no momentum. Weight clipping has since been abandoned in the favour of other methods, but as we show in Figure~\ref{fig:sn_mom} and Appendix~\ref{app:exp} current methods like Spectral Normalization
applied to GAN critics trained with low momentum decrease sensitivity to learning rates but perform poorly in conjunction with high momentum, leading to slower convergence and higher hyperparameter sensitivity.

We have shown that smoothness constraints interact with optimization parameters such as learning rates and momentum, and argue that we need to reassess our understanding of smoothness constraints, not only as constraints on the final model, but as methods which influence the \textit{optimization path}.

\textbf{Sensitivity to data scaling}. Sensitivity to data scaling of smoothness constraints can
make training neural network models sensitive to additional hyperparameters.
Let $f^*$ be the optimal decision function for a task obtained from using i.i.d samples from random variable $X$, and
$f_c^*$ obtained similarly from i.i.d samples obtained from $c X$.
Since $f^*$ and $f_c^*$ can be highly non linear, the relationship between the smoothness of the two functions is unclear.
 This gets further complicated when we consider their closest approximators under a neural family.
The effect of data scaling on the smooth constraints
 required to fit a model can be exemplified using the two moons dataset:
 with a Lipschitz constraint of 1 on the model the data is poorly fit
 - Figure~\ref{fig:two_moons_decision_surface_sn} -
but a much better fit can be obtained by changing the Lipschitz constant to 10 - Figure~\ref{fig:two_moons_decision_surface_sn_10} -
or scaling the data - Figure~\ref{fig:two_moons_decision_surface_sn_data_scaling}.

\textbf{Wrong model priors}. The wrong kind of smooth functions can have a similar effect to restricting capacity by introducing the wrong inductive biases. For many regularization techniques it is unclear \textit{what kind} of smoothness they are encouraging and how strong their effect is on the smoothness of the learned function.
Lipschitz smoothness constraints on image models are often specified with respect to the
 $l_2$ norm, which is notorious for not being a meaningful distance metric for natural images.
Why are we using the power of feature learning if we are restricting our models to be part of a family of functions which impose constraints that rely on rigid distance metrics? We have seen that smooth critics can be a catalyst for learning by providing the right signal, but that is only if their similarity measures are relevant for the task.

\section{Paving the path towards smoothness in neural networks}

Smoothness regularization of neural networks has brought forward advances in a plethora of machine learning tasks,
from supervised to unsupervised and reinforcement learning.
These advances are just scratching the surface of the benefits of smoothness, to explore its full potential we have to use new smoothness definitions, distance metrics and complexity measures;
we need to define task and modality dependent smoothness constraints applied to learned representations;
we have to adopt an integrated view and understand its interactions with
losses, data, model architectures and optimization.
Meanwhile,
in the world of task agnostic smoothness applied to high dimensional data spaces,
we continue to be surprised that specifying smoothness constraints is not better.

\textbf{New ways of defining smoothness}. Improving model generalization and robustness
requires specifying the right level of invariance by using task information to define smoothness constraints.
To solve issues in generative modelling such as ``mode dropping'' or ``mode collapse'',
where entire modes from the data distribution are not captured by the model, we have to go beyond
current smoothness measures such as Lipschitz continuity defined on the input space.
With the right feature space, images of cats are close to images of dogs, and thus a model which is smooth in that space is less likely to drop one of the two modes.
We have to ask what are the desired properties of $h$ such that $\norm{f(h(\vx)) - f(h(\vy))}  \le \norm{h(\vx) - h(\vy)}$ instead of applying the smoothness constraints on the raw data.
Since we require that the mapping $h$ does not discard task relevant information in the data, maintains useful diversity and accounts for input modalities, it has to be \textit{data} and \textit{task} dependent.
As we have seen again and again in the development of machine learning, handcrafting $h$ is not a scalable solution, and thus the mapping itself has to be \textit{learned}.
Since the approaches used for learning the right representations have to be task dependent, different insights will
be required for supervised and unsupervised methods.
We expect that semi-supervised learning will play an important role, and that hints
for useful properties of representation domains will come from representation learning methods and inference techniques.

\textbf{New ways of measuring smoothness}. Measuring smoothness of a function parametrized by a neural network is challenging even for the most common measure of smoothness used in machine learning, Lipschitzness.
Loose upper and lower bounds which rely on function composition are often used~\citep{virmaux2018lipschitz,combettes2019lipschitz}.
~\citet{fazlyab2019efficient} provide an algorithm with tighter bounds by leveraging that activation
functions are derivatives of convex functions and cast finding the Lipschitz constant as the result of a convex
optimization problem.
However, their most accurate approach scales quadratically with the number of neurons and only applies to feed forward networks.
~\citet{sokolic2017robust} provide an upper bound for the Lipschitz constant of a neural network with linear,
softmax and pooling layers restricted to input space $\mathcal{X}$ via $\norm{f(\vx) - f(\vy)}_{2} \le \sup_{\vz \in \text{convex\_hull}(\mathcal{X})} J(\vz) \norm{\vx-\vy}_2,  \hspace{1em} \forall \vx, \vy \in \mathcal{X}$,
but empirically resort to layerwise bounds. If we want to understand the effects of network architectures,
regularization methods and optimization algorithms have on model smoothness,
we have to be able to accurately measure it.

\textbf{New learning paradigms}.
Combining non parametric methods with feature learning is a promising approach to learning smooth decision surfaces.
Given the right data representations and the appropriate distance metrics,
interpolating between training examples is an excellent and interpretable smoothness prior.
Pursuing this avenue of research entails learning the right features, which themselves might have to be smooth~\citep{balaji_uncertanty_smoothness}, as well as further avenues for scaling non parametric methods such as Gaussian Processes, Support Vector Machines and Nearest Neighbours methods to large datasets.

\textbf{New measures of model complexity}.
Standard complexity measures, from VC dimensions and Rademacher complexity, to simpler measures such as number of learned parameters ignore the problem the model has to solve.
Task definitions need to be accounted for in the new generation of model complexity measures, since fitting
random labels (as per Rademacher complexity) discounts the inductive bias in smoothness constraints that can help model fitting and generalization.
The issue of measuring model complexity is inherently tied with many other issues discussed so far,
 such as choosing ways to define and quantify smoothness.

\textbf{New approaches to old problems}. Smooth learned critics have advanced the state of art in generative modeling and reinforcement learning.
Why stop there? By viewing parametric critics as learned loss functions, and observing
 that for any lower layer of a neural network the upper layers are part of a learned loss function,
we can further explore and expand the benefits of smoothness.
By exploring how smoothness helps critics in non stationary environments such as reinforcement learning and generative models, we can solve notorious neural network training problems such as covariate shift.

\clearpage

\textbf{Acknowledgments}. We would like to thank Suman Ravuri for helpful comments and discussions.
\bibliography{references}

\clearpage
\appendix
\input{appendix}

\end{document}

%% file: macros.tex
\usepackage{amsmath, latexsym}
\usepackage{amsfonts}
\usepackage{mathtools}
\usepackage{ntheorem}
\usepackage{dsfont}
\usepackage[dvipsnames]{xcolor}
\usepackage[colorinlistoftodos]{todonotes}
\usepackage{booktabs}
\usepackage{xfrac}
\usepackage{bbm}

\usepackage{algorithm}
\usepackage{algorithmicx}

\usepackage[most]{tcolorbox}
\usepackage{xparse}
\usepackage{lipsum}
\usepackage{changepage}
\usepackage{enumitem}

\newcommand{\qedwhite}{\hfill \ensuremath{\Box}}

\newcommand\cut[1]{}

\usepackage[percent]{overpic}
\newcommand{\norm}[1]{\left\lVert#1\right\rVert}

\newcommand{\squishlist}{
   \begin{list}{$\bullet$}
    { \setlength{\itemsep}{0pt}      \setlength{\parsep}{3pt}
      \setlength{\topsep}{3pt}       \setlength{\partopsep}{0pt}
      \setlength{\leftmargin}{1.5em} \setlength{\labelwidth}{1em}
      \setlength{\labelsep}{0.5em} } }

\newcommand{\squishlisttwo}{
   \begin{list}{$\bullet$}
    { \setlength{\itemsep}{0pt}    \setlength{\parsep}{0pt}
      \setlength{\topsep}{0pt}     \setlength{\partopsep}{0pt}
      \setlength{\leftmargin}{2em} \setlength{\labelwidth}{1.5em}
      \setlength{\labelsep}{0.5em} } }

\newcommand{\squishend}{
    \end{list}  }

\newcommand{\myvec}[1]{\mathbf{#1}}
\newcommand{\myvecsym}[1]{\boldsymbol{#1}}

\newcommand{\vtheta}{\myvecsym{\theta}}

\newcommand{\vx}{\myvec{x}}

\newcommand{\vy}{\myvec{y}}

\newcommand{\vz}{\myvec{z}}

\newcommand{\be}{\begin{equation}}
\newcommand{\ee}{\end{equation}}
\newcommand{\bea}{\begin{eqnarray}}
\newcommand{\eea}{\end{eqnarray}}
\newcommand{\beaa}{\begin{eqnarray*}}
\newcommand{\eeaa}{\end{eqnarray*}}

\DeclareMathAlphabet{\mathpzc}{OT1}{pzc}{m}{n}

%% file: appendix.tex
\section{Appendix}

\subsection{Additional experimental results and experimental methodology}
\label{app:exp}

\textbf{Architectures}. In all the two moons experiments, the DeepMLP has 4 layers and 100, 100, 100 and 1 output units respectively. The shallow MLP has 2 layers of 100 and 1 unit. All methods were trained for 100 iterations on 50 datapoints. The MNIST plots are obtained from MLP classifiers having 4 layers of 1000, 1000, 1000 and 10 units each
and are trained for 500 iterations at batch size 100, reaching an accuracy of 95\% on the entire test set.
For the GAN CIFAR-10 experiments, we use the architectures specified in the Spectral normalization paper~\citep{miyato2018spectral}.
Unless otherwise specified, we use the default Adam optimizer~\citep{kingma2014adam} $\beta_1$ and $\beta_2$ parameters.

\textbf{Computing the local Lipschitz constant in Figure~\ref{fig:two_moons_layers_local_constant}}.
To compute the local Lipschitz function of the decision surface learned on two moons, we split the space into small neighborhoods (2500 equally sized grids). For each grid, we sample 2500 random pairs of points in the grid and report $\max \norm {f(\vx) - f(\vy)} / \norm{\vx - \vy}$.

\begin{figure}[h]
  \centering
  \begin{subfigure}[b]{0.32\textwidth}
  \includegraphics[width=\textwidth]{sn_momentum_0_5}
  \caption{Low momentum: 0.5.\newline}
  \end{subfigure}
  \begin{subfigure}[b]{0.32\textwidth}
  \includegraphics[width=\textwidth]{sn_momentum_0_9}
  \caption{High momentum: 0.9.\newline}
  \end{subfigure}
  \begin{subfigure}[b]{0.32\textwidth}
  \includegraphics[width=\textwidth]{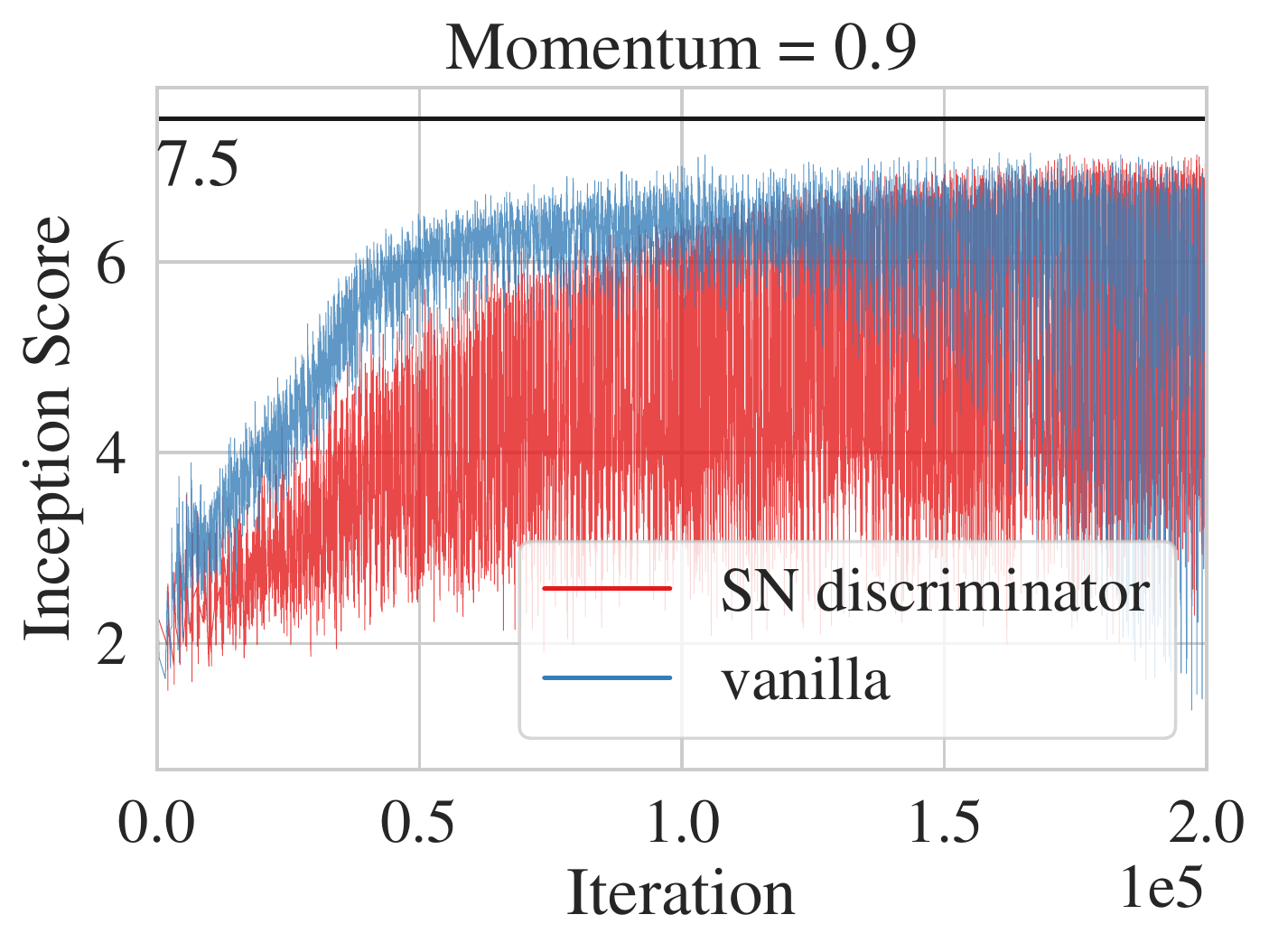}
  \caption{High momentum with 10 power iteration caching and no caching.}
  \end{subfigure}
  \caption{The effect of momentum on spectral normalization on GAN performance. This shows that the iteration between momentum and spectral normalization is not due to the caching between iterations done for computational reasons.}
  \label{fig:sn_mom_with_no_caching}
\end{figure}

\textbf{Spectral normalization}.
In Figure~\ref{fig:sn_mom_with_no_caching} we show that the effect of momentum on spectral normalization is independent of whether caching of the initialization vector for power iteration is performed or not.

